\documentclass[10pt,twocolumn,letterpaper]{article}

\usepackage{cvpr}              

\usepackage[dvipsnames]{xcolor}

\usepackage{booktabs}
\usepackage{adjustbox}
\usepackage{color}
\usepackage[dvipsnames]{xcolor}

\definecolor{cvprblue}{rgb}{0.21,0.49,0.74}
\usepackage[pagebackref,breaklinks,colorlinks,citecolor=cvprblue]{hyperref}

\def\paperID{304} 
\def\confName{3DV\xspace}
\def\confYear{2026\xspace}

\title{Structural Energy-Guided Sampling for View-Consistent Text-to-3D}

\author{Qing Zhang$^{1}$,  Jinguang Tong$^{1,2}$, Jie Hong$^{3}$, Jing Zhang$^{1}$, Xuesong Li$^{1, 2}$
\\
\small $^{1}$ The Australian National University \quad $^{2}$CSIRO, \quad $^{3}$ The University of Hong Kong\\
}

\begin{document}
\maketitle
\begin{abstract}
Text-to-3D generation often suffers from the Janus problem, where objects look correct from the front but collapse into duplicated or distorted geometry from other angles. We attribute this failure to viewpoint bias in 2D diffusion priors, which propagates into 3D optimization.
To address this, we propose Structural Energy-Guided Sampling (SEGS), a training-free, plug-and-play framework that enforces multi-view consistency entirely at sampling time. SEGS defines a structural energy in a PCA subspace of intermediate U-Net features and injects its gradients into the denoising trajectory, steering geometry toward the intended viewpoint while preserving appearance fidelity.
Integrated seamlessly into SDS/VSD pipelines, SEGS significantly reduces Janus artifacts, achieving improved geometric alignment and viewpoint consistency without retraining or weight modification.
\end{abstract}    
\section{Introduction}
\label{sec:intro}
High-quality 3D assets are foundational to entertainment, product design, AR/VR, and simulation. Despite the rapid progress of text-to-image generation driven by large-scale web datasets~\cite{schuhmann2022laion}, the scarcity and cost of broad-coverage 3D supervision~\cite{deitke2024objaverse} still constrain text-to-3D pipelines. This raises a central question: how can we transfer the rich priors of powerful 2D diffusion models into reliable, multi-view consistent 3D representations without relying on large curated 3D corpora? A seminal step in this direction is DreamFusion~\cite{poole2022dreamfusion}, which optimizes a Neural Radiance Field (NeRF) from a frozen text-to-image diffusion prior via Score Distillation Sampling (SDS). The SDS paradigm has since inspired numerous systems~\cite{chen2023fantasia3d, lin2023magic3d, metzer2023latent, liang2024luciddreamer, zhang2024viewpoint, han2024latent, wang2024prolificdreamer, cao2024dreamavatar}, making high-fidelity text-conditioned synthesis feasible without paired 3D training data.

\begin{figure}[t]
    \centering
       \includegraphics[width=\linewidth]{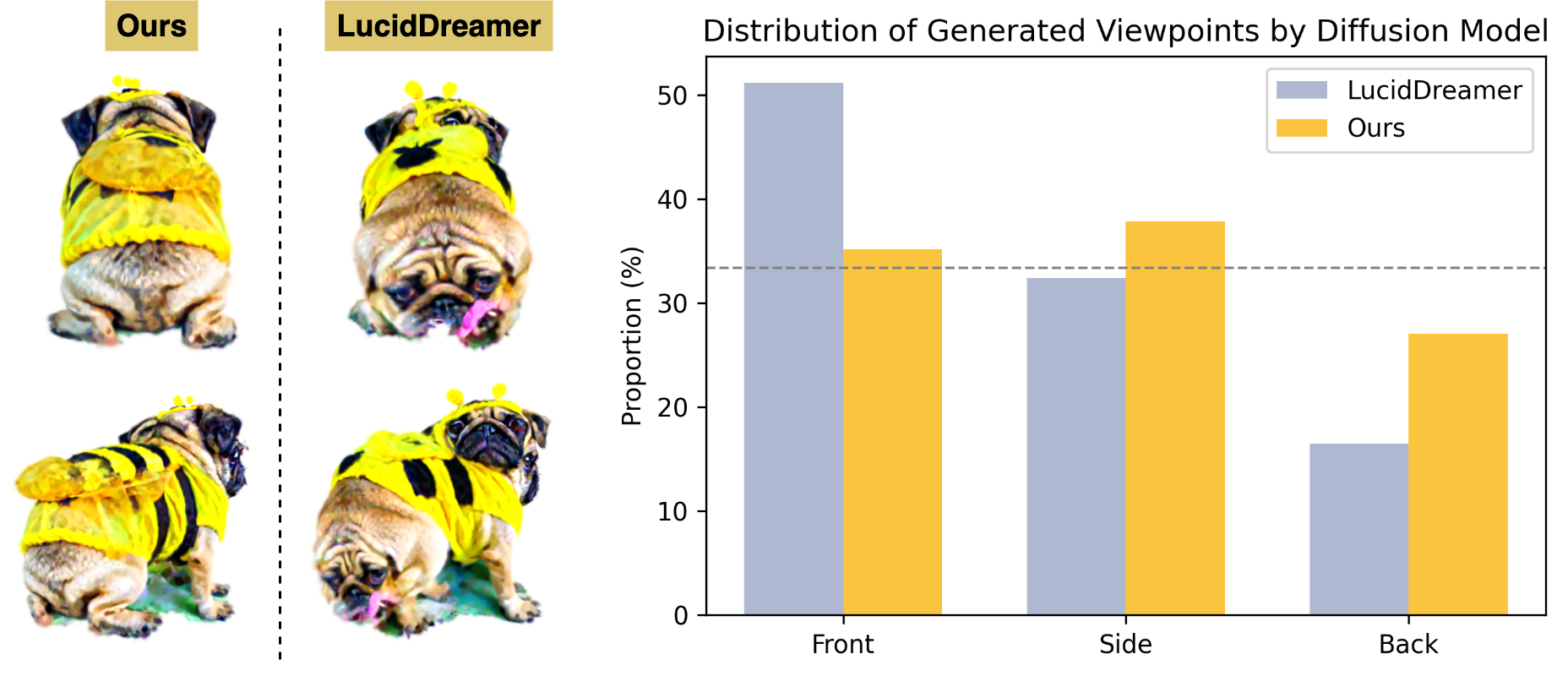}
    \caption{
        Comparison of view distributions generated by LucidDreamer~\cite{liang2024luciddreamer} and our method, using the prompt “a DSLR photo of a pug wearing a bee costume.” Our method yields a more uniform view sampling distribution, whereas LucidDreamer exhibits a typical long-tail pattern. As a result, the multi-head issue is significantly alleviated in our outputs.}
    \label{fig:example_motivation}
\end{figure}


Despite the practical success of SDS-style pipelines, they frequently suffer from the Janus problem: shapes that appear plausible from a canonical front view deteriorate into duplicated or distorted geometry from other angles (Fig.~\ref{fig:example_motivation}). One line of remedies replaces the frozen 2D prior with a multi-view–adapted version, by retraining or fine-tuning on curated view supervision, as in Zero-1-to-3 and MVDream~\cite{liu2023zero, shi2023mvdream}. Although this improves pose coverage, updating the score network reshapes the appearance manifold: high-frequency textures and material cues are weakened, styles drift toward the adapted domain, and cross-category generalization declines~\cite{huang2024dreamcontrol}. Moreover, the supervision, curation, and compute overhead reduce the plug-and-play appeal of SDS/VSD. Other attempts instead steer generation with external proxies such as edges, depth, sketches, or layout, using control branches or multi-view conditioning~\cite{huang2024dreamcontrol}.

\begin{figure}[t]
    \centering
\includegraphics[width=\linewidth,height=0.35\textheight,keepaspectratio]{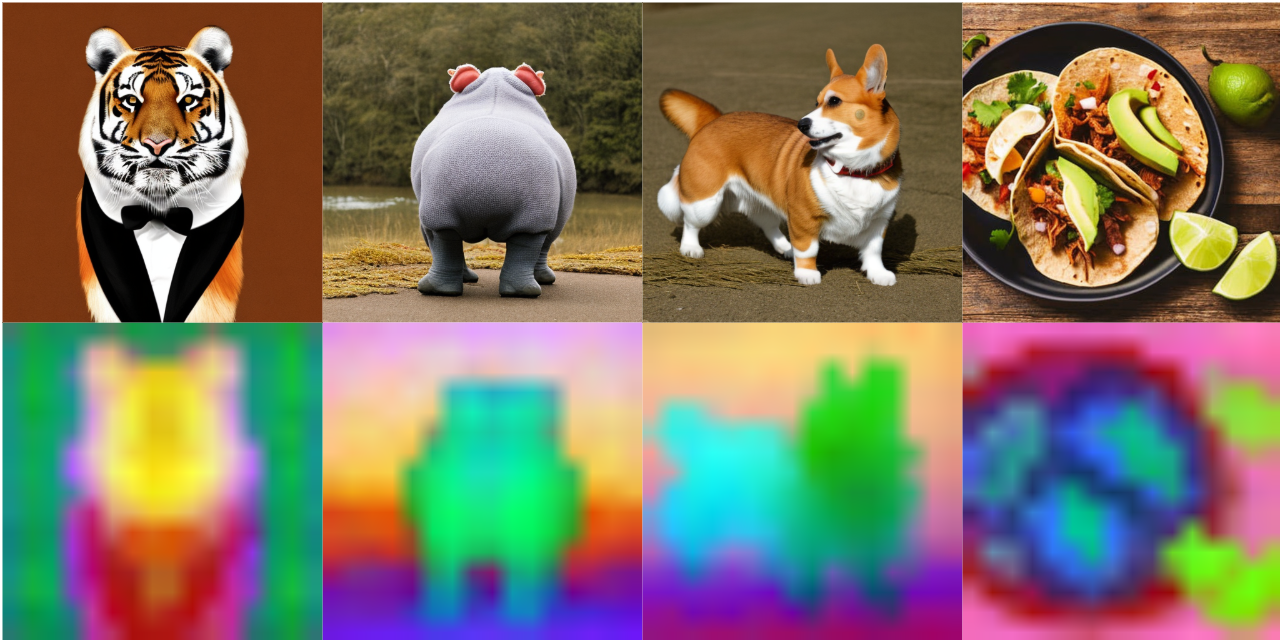}
    \caption{PCA of the top–3 components from the first decoder self-attention module, highlighting object silhouettes and part boundaries, showing early decoder features encode stable structural cues}
    \label{fig:pca_visualization}
\end{figure}

Text-to-image diffusion priors trained on Internet photos exhibit a viewpoint bias toward frontal views. SDS optimizes the 3D representation so that renders from random viewpoints are consistently scored as high-likelihood images by the frozen prior. This process unintentionally reinforces the prior’s frontal preference across all poses, ultimately leading to Janus artifacts. 
Our key insight is that the pretrained diffusion U-Net already encodes rich, view-relevant structural information in its early and intermediate layers.
As shown in Fig.~\ref{fig:pca_visualization}, a simple PCA visualization of early decoder self-attention features reveals stable object silhouettes and part boundaries. 
If such a structure is exposed as an energy-like target and allowed to steer the sampling trajectory, the generation can be guided toward the intended structural configuration of the target view, counteracting the bias while preserving fine textures and material details.

\begin{figure}[t]
    \centering
    \includegraphics[width=\linewidth,height=0.35\textheight,keepaspectratio]{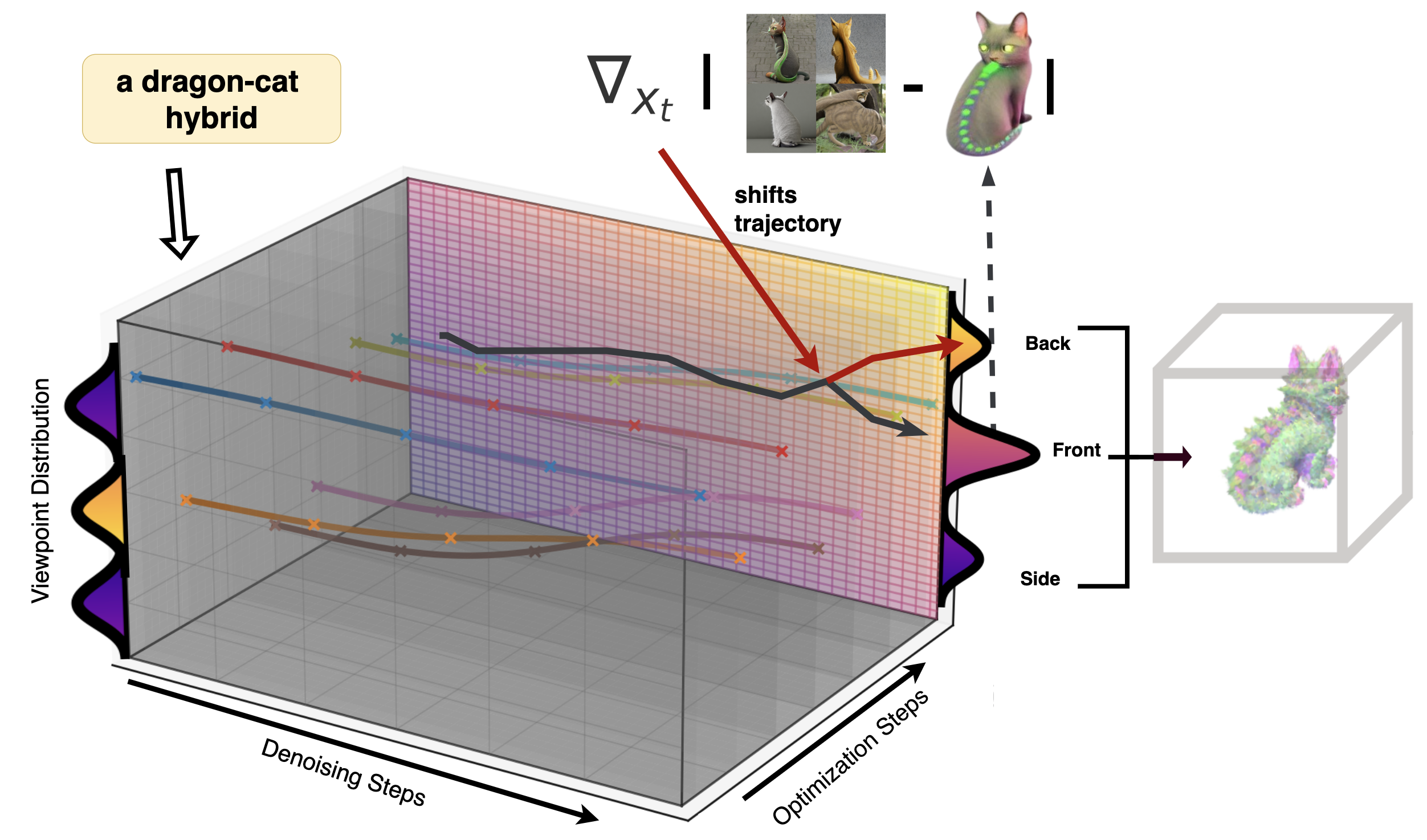}
    \caption{Illustration of our motivation. The entire text-to-3D optimization can be interpreted as multiple denoising trajectories evolving in latent space. The vertical axis represents the viewpoint distribution. During the sampling phase, viewpoints (front, side, back) are drawn uniformly, yet the diffusion prior tends to bias trajectories toward frontal views. Our key idea is to exploit structural features extracted from underrepresented back-view samples to inject corrective gradients, thereby shifting trajectories toward the desired back viewpoint.}
    \label{fig:motivation}
\end{figure}

To address this issue, we propose \textbf{SEGS} (Structural Energy-Guided Sampling), a training-free approach that improves multi-view consistency during generation while preserving fine appearance. SEGS offers two complementary signals: \textit{Structural Energy Guidance}, which exposes the structure encoded in intermediate U\hbox{-}Net activations by constructing a viewpoint-aware PCA subspace and defining a simple structural energy whose gradient is injected into the denoising update to steer the sampling trajectory toward the target view while keeping the diffusion weights frozen(Fig.~\ref{fig:motivation}); and a lightweight \textit{Text Consistency Guard}, which optionally prunes supervision that is clearly inconsistent with the requested viewpoint to stabilize edge cases. Operating purely at sampling time, SEGS integrates seamlessly with SDS/VSD and controls the trajectory rather than the parameters.
In summary, our contributions are as follows:
\begin{itemize}
\item We introduce \textbf{SEGS}, a sampling-time formulation that casts view consistency as minimizing a structural energy in a PCA subspace of intermediate diffusion features, steering geometry toward the target view without changing model weights.
\item The approach is \textbf{training-free} and \textbf{plug-and-play}, integrates with SDS/VSD, and \textbf{no additional predictors}, preserving high-frequency textures and material fidelity.
\item Across strong text-to-3D baselines, SEGS reduces Janus Problem and improves multi-view alignment under the same compute budget.
\end{itemize}

\begin{figure*}[htbp]
    \centering
    \includegraphics[width=1.0\linewidth]{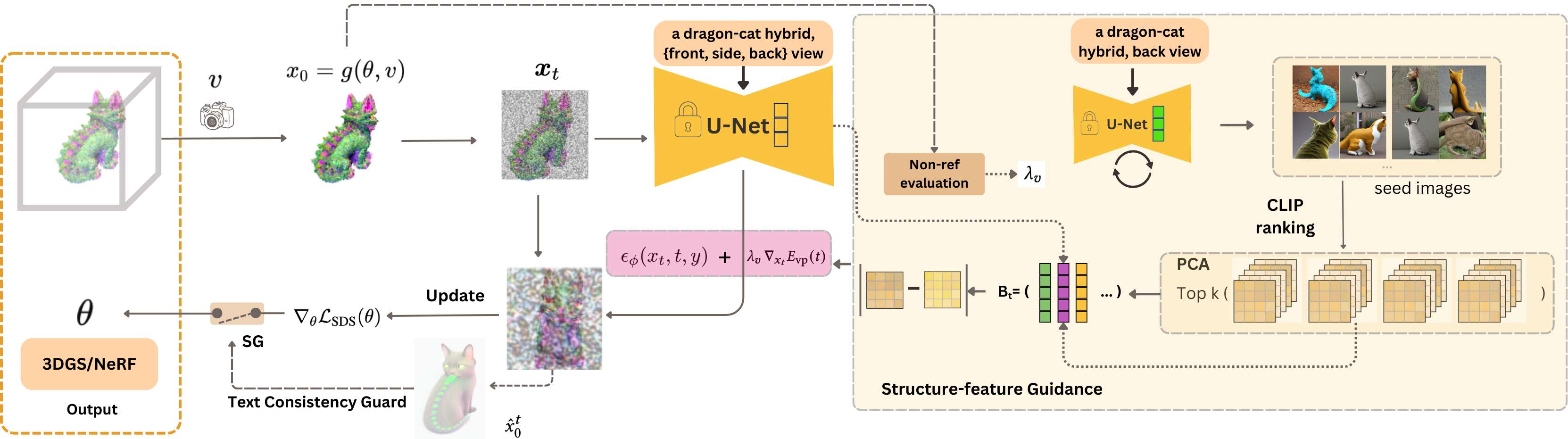} 
    \caption{Overview of our pipeline. We formulate view consistency as \emph{energy-guided sampling}: a PCA-based structural subspace is constructed from intermediate U-Net features to define a viewpoint-aware \emph{structural energy}. Its gradient is injected into the noise prediction at each denoising step, steering the trajectory toward the target viewpoint \emph{without} training or weight updates. An optional text-consistency guard can prune misaligned supervision.}
    \label{fig:pipeline}
\end{figure*}

\section{Related Work}
\label{sec:related_work}

\subsection{Paradigms of Text-to-3D Generation}
Existing text-to-3D frameworks broadly fall into three paradigms: end-to-end generation, SDS-based 2D-to-3D distillation, and reconstruction-based generation.

\noindent \textbf{End-to-End Generation.} 
Early methods trained a single network to directly map text embeddings into 3D representations, such as Point-E~\cite{nichol2022point} and Shap-E~\cite{jun2023shap}. While conceptually simple and fast at inference, these approaches remain limited in fidelity and category coverage due to the scarcity of large-scale 3D data~\cite{liu2024comprehensive}, motivating the shift toward leveraging 2D priors.

\noindent \textbf{SDS-based 2D-to-3D Distillation.} 
DreamFusion~\cite{poole2022dreamfusion} introduced Score Distillation Sampling (SDS), which quickly became the dominant paradigm by using frozen 2D diffusion priors to optimize NeRFs~\cite{mildenhall2021nerf, xiao2025neural}. 
Subsequent works improved quality (e.g., ProlificDreamer~\cite{wang2024prolificdreamer}, Fantasia3D~\cite{chen2023fantasia3d}) or speed, such as GaussianDreamer~\cite{yi2024gaussiandreamer} and DreamGaussian~\cite{tang2023dreamgaussian}, which exploits Gaussian Splatting for fast rendering~\cite{kerbl20233d, tong2025gs, li2024dgns}. However, because they inherit viewpoint biases from the underlying 2D models, these pipelines still suffer from Janus artifacts.

\noindent \textbf{Reconstruction-based Generation.} 
Recent methods such as Instant3D~\cite{li2023instant3d} and LGM~\cite{tang2024lgm} regress 3D assets from sparse multi-view images synthesized by fine-tuned diffusion models. These approaches are efficient and high-resolution, yet still rely on biased 2D priors, leading to similar multi-view inconsistencies. 

\subsection{View Consistency in Text-to-3D}
Efforts to address view inconsistency fall into three directions: retraining diffusion priors, injecting external supervision, and sampling-time guidance.

\noindent \textbf{Retraining or Fine-tuning.}
Methods such as Zero-1-to-3~\cite{liu2023zero} and MVDream~\cite{shi2023mvdream} enhance multi-view awareness by fine-tuning diffusion models on curated multi-view datasets with 3D-aware attention. While effective, these approaches rely on domain-specific 3D corpora—often stylized or synthetic—which leads to weakened textures, reduced detail fidelity, and poorer cross-category generalization~\cite{huang2024dreamcontrol}. They also require additional compute, undermining the plug-and-play appeal of SDS pipelines.

\noindent \textbf{External Supervision.}
Another line introduces proxy signals such as edges, depth, or sketches through ControlNet-style conditioning~\cite{huang2024dreamcontrol, chen2023control3d}. These methods stabilize geometry but demand extra predictors and training, and they do not provide explicit structural targets for viewpoint alignment.

\noindent \textbf{Sampling-time Guidance.}
A third direction keeps the backbone frozen and modifies the guidance signal at inference. For example, D-SDS~\cite{hong2023debiasing} debiases the score and prompt, and Perp-Neg~\cite{armandpour2023re} reparameterizes negative prompts to mitigate semantic conflicts. While lightweight and plug-and-play, such methods operate only in text/score space, without offering structural cues.
\section{Preliminary}
\label{sec:pre}

\subsection{SDS for Text-to-3D }
Let $x_0=g(\theta,v)$ denote the rendered image of a differentiable 3D representation with parameters $\theta$ under camera view $c$. Let $\epsilon_\phi(x_t,y)$ be the noise prediction of a pretrained text-to-image diffusion model, conditioned on the noisy image $x_t$, timestep $t$, and text prompt $y$. The Score Distillation Sampling (SDS) gradient~\cite{poole2022dreamfusion} is
\begin{equation}
\nabla_{\theta} \mathcal{L}_{\text{SDS}}(\theta) \approx 
\mathbb{E}_{t,\epsilon,c} \!\left[ \,\omega(t)\,\big(\epsilon_{\phi}(x_t, y) - \epsilon_t\big)\,\frac{\partial x_0}{\partial \theta} \right],
\label{equation1}
\end{equation}
where $\epsilon_t$ is Gaussian noise and $\omega(t)$ is a timestep weighting.
From DDPM~\cite{ho2020denoising,song2020denoising}, the clean-image estimate is
\begin{equation}
 \hat{x}_0^t\ = \frac{x_t - \sqrt{1 - \bar{\alpha}_t}\,\epsilon_{\phi}(x_t, y)}{\sqrt{\bar{\alpha}_t}},
 \label{equation8}
\end{equation}
with $\bar{\alpha}_t=\prod_{i=1}^{t}\alpha_i$ and $\gamma(t)=\sqrt{(1-\bar{\alpha}_t)/\bar{\alpha}_t}$. An equivalent form of Eq.~(\ref{equation1}) should be Eq.(~\ref{equation9}),
\begin{equation}
\nabla_{\theta} \mathcal{L}_{\text{SDS}}(\theta) =
\mathbb{E}_{t,\epsilon,c} \!\left[ \frac{\omega(t)}{\gamma(t)}\,\big(x_0-\hat{x}_0^t\big)\,\frac{\partial x_0}{\partial \theta} \right].
\label{equation9}
\end{equation}
This formulation provides a standard interface for modifying the denoising prediction used by SDS, which we will exploit to introduce a structure-aware trajectory correction.

\subsection{Energy-Guided Denoising}
Following the general principle of energy-based guidance in diffusion sampling~\cite{dhariwal2021diffusion,epstein2023diffusion}, we adopt an energy-based view of controllable sampling. At each denoising step, an additional, task-specific directional term derived from a differentiable \emph{energy} is injected into the diffusion model’s noise prediction to steer the sampling trajectory while keeping the backbone frozen:
\begin{equation}
\hat{\epsilon}_\phi(x_t,y)\;=\;\epsilon_\phi(x_t,y)\;+\;\lambda_v(t)\,\nabla_{x_t}E(x_t,t),
\label{eq:eg_overview}
\end{equation}
where $E(x_t,t)$ is designed such that lower values correspond to more desirable samples, and $\lambda_v(t)$ controls the guidance schedule.
Note that since DDPM/DDIM updates subtract $\hat{\epsilon}_\phi$ from the state, the $+\,\nabla_{x_t}E$ term in Eq.~(\ref{eq:eg_overview}) induces a descent step on $E$ at the state level.
This yields a training-free, plug-and-play mechanism to reshape the denoising path. In Section~\ref{sec:method}, we instantiate $E$ as a \emph{PCA-structural energy} defined in a viewpoint-aware structural subspace and detail its coupling to the SDS objective.

\subsection{Viewpoint Bias in SDS Guidance}
Although SDS provides a convenient interface for supervision (Eq.~\ref{equation9}), the quality of its guidance depends entirely on the frozen diffusion prior. 
In practice, diffusion models trained on Internet photos exhibit a strong bias toward frontal viewpoints. 
As a result, the pseudo-targets $\hat{x}_0^t$ generated during optimization are not uniformly distributed across poses: frontal predictions dominate even when camera poses are sampled uniformly(Fig~\ref{fig:motivation}). 

To verify this, we log $\hat{x}_0^t$ throughout the SDS optimization and classify their viewpoints (front/side/back) with a CLIP-based classifier. 
The resulting histogram (Fig.~\ref{fig:example_motivation}) reveals a clear imbalance: frontal views are over-represented, while side and back views are significantly under-sampled. 
This mismatch distorts the gradients received by non-frontal views, reinforcing the prior’s bias at every optimization step and manifesting as Janus artifacts. 

Taken together, these observations highlight a fundamental limitation of SDS: it provides \emph{imbalanced structural guidance}. 
This motivates our approach to explicitly expose and leverage structural cues from intermediate diffusion features, counteracting viewpoint bias without modifying the diffusion weights.


\section{Method}
\label{sec:method}

As outlined in Section~\ref{sec:pre}, we adopt an \emph{energy-guided denoising} perspective, casting view-consistency preservation as minimization of a viewpoint-aware \emph{structural energy} within the iterative denoising process. 
Rather than retraining the backbone diffusion model or rebalancing its data distribution, we introduce a training-free, plug-and-play mechanism that steers each sampling step toward geometry consistent with the target viewpoint.

Self-attention modules in diffusion models contain rich structural representations, capturing spatial layout, object pose, and geometry across different generation stages~\cite{mo2024freecontrol}. 
Beyond their internal role in denoising, these intermediate features can be extracted and repurposed as external guidance signals to influence the sampling trajectory~\cite{epstein2023diffusion}. 
Motivated by this insight, we extract such features from the self-attention blocks within the U\mbox{-}Net decoder and project them into a viewpoint-aware PCA subspace. 
By defining a structural energy in this subspace and steering the denoising process toward features from viewpoint-specific exemplars, our framework provides a training-free, geometry-aware mechanism that improves view consistency in text-to-3D generation. 
A lightweight CLIP-based semantic guard can optionally be applied to prune clearly misaligned supervision; its role is auxiliary and details are deferred to the supplement.

This section details the construction of the PCA-based structural subspace (Sec.~\ref{sec:struct_pca}), the definition of viewpoint-aware structural energy (Sec.~\ref{sec:struct_energy}), its integration into the denoising update (Sec.~\ref{sec:eg_update}), and the coupling with SDS/VSD (Sec.~\ref{sec:coupling}).

\subsection{Structural Subspace via PCA}
\label{sec:struct_pca}

\noindent
We construct a viewpoint-aware structural subspace from intermediate self-attention features of the diffusion U\mbox{-}Net and use it as the basis for subsequent energy definition and guidance.

\paragraph{Sampling structural features.}
\label{samplig_features}
Given a text prompt $y$ augmented with a target-view token (e.g., \emph{``back view''}), we first generate $N$ auxiliary images using the pretrained diffusion model.
Following~\cite{mo2024freecontrol}, for the $i$-th image at denoising timestep $t$, we extract self-attention \emph{keys} $\mathbf{b}_{t,i}\!\in\!\mathbb{R}^{H\times W\times C}$ from the final layer of the first decoder up\_block of the U\mbox{-}Net (unless otherwise stated).
Such features are known to encode spatially aligned representations~\cite{tumanyan2023plug}.
Collecting across $N$ samples yields a tensor $\mathbf{b}_{t}\!\in\!\mathbb{R}^{N\times H\times W\times C}$.

We then apply Principal Component Analysis (PCA) to $\mathbf{b}_{t}$ (after channel-wise mean centering) to identify dominant directions and form a common feature subspace.
Retaining the first $N_b$ principal components gives
\begin{equation}
\mathbf{B}_t = \text{PCA}(\mathbf{b}_{t}),
\end{equation}
where $\mathbf{B}_t \in \mathbb{R}^{C \times N_b}$.
Unless otherwise noted, $\mathbf{B}_t$ is built at the \emph{current} denoising timestep $t$ to ensure temporal alignment.

Independently, we compute CLIP~\cite{radford2021learning} similarities between each generated image and the target-view text, and select the top-$k$ images.
The corresponding features are aggregated as $\mathbf{F}_t \in \mathbb{R}^{k \times H \times W \times C}$ and projected into the PCA subspace:
\begin{equation}
\mathbf{S}_{g,t} = \mathbf{F}_t \mathbf{B}_t,
\label{eq:struture-guided-feature}
\end{equation}
yielding $\mathbf{S}_{g,t} \in \mathbb{R}^{k \times H \times W \times N_b}$ as viewpoint-aware structural references distilled from the auxiliary set.
During optimization, we treat $\mathbf{B}_t$ and $\mathbf{S}_{g,t}$ as constants (no gradient flows into the PCA pipeline).

\begin{figure}[t]
    \centering
    \includegraphics[width=1.\linewidth]{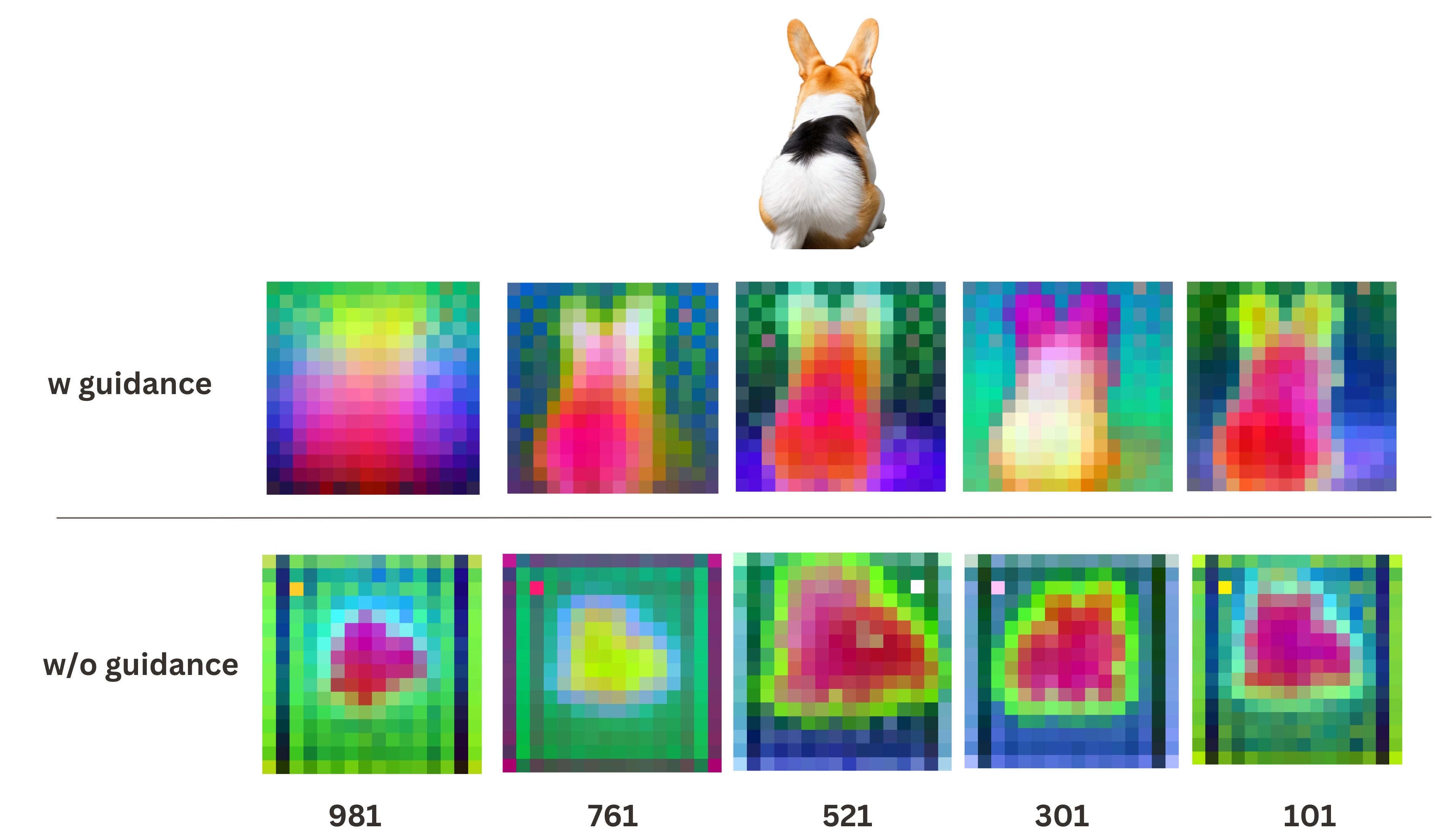}
    \caption{Feature maps with and without structural guidance. We perform PCA-based dimensionality reduction on intermediate self-attention features of the U-Net. The top row shows an image with a back view; the middle row visualizes intermediate features with structure-feature guidance; the bottom row shows features without structural supervision.}
    \label{fig:generation_bais}
\end{figure}

\subsection{Viewpoint Target and Structural Energy}
\label{sec:struct_energy}

\noindent
With the PCA structural basis $\mathbf{B}_t$ and the reference set $\mathbf{S}_{g,t}$ prepared in Sec.~\ref{sec:struct_pca}, we now define a structural energy that quantitatively measures alignment to the target-view structure.

\paragraph{Projected current feature.}
At each text-to-3D iteration (Fig.~\ref{fig:pipeline}), given the noisy input $x_t$, we extract the U\mbox{-}Net intermediate self-attention feature $\mathbf{f} \in \mathbb{R}^{H \times W \times C}$ and project it onto $\mathbf{B}_t$ (projection along the channel dimension):
\begin{equation}
\mathbf{G}_t \;=\; \mathbf{f}\,\mathbf{B}_t,
\end{equation}
where $\mathbf{G}_t \in \mathbb{R}^{H \times W \times N_b}$ is the subspace-projected representation of the current state.

\paragraph{Structural energy.}
To guide $\mathbf{G}_t$ toward the target-view structural references $\mathbf{S}_{g,t}$ (top-$k$ by CLIP similarity), we define the viewpoint guidance energy $E_{\text{vp}}(t)$ as the average mean squared error between $\mathbf{G}_t$ and each reference $\mathbf{S}_{g,t}[n]$:
\begin{equation}
\begin{adjustbox}{width=0.48\textwidth}
$E_{\text{vp}}(t) 
= \frac{1}{k} \sum_{n=1}^{k} 
\frac{1}{H \times W \times N_b} 
\sum_{i=1}^{H \times W} \sum_{j=1}^{N_b}
\left(\mathbf{G}_t[i,j] - \mathbf{S}_{g,t}[n,i,j]\right)^2$
\end{adjustbox}
\end{equation}
Here, $i$ and $j$ index the flattened spatial location and PCA-reduced channel, respectively, and $n=1,\dots,k$ indexes the selected references.
Lower $E_{\text{vp}}(t)$ indicates better structural consistency.
Gradients with respect to $x_t$ are computed via the chain rule $\nabla_{x_t}E_{\text{vp}}(t)=\frac{\partial E}{\partial \mathbf f}\frac{\partial \mathbf f}{\partial x_t}$ while $\mathbf{B}_t$ and $\mathbf{S}_{g,t}$ remain fixed.

\subsection{Energy-Guided Denoising Update}
\label{sec:eg_update}

\noindent
With $\mathbf{B}_t$, $\mathbf{S}_{g,t}$, and $E_{\text{vp}}(t)$ defined in Sec.~\ref{sec:struct_energy}, we inject the corresponding gradient into the denoising prediction to steer the sampling trajectory toward the target viewpoint while keeping the diffusion backbone frozen.
At each denoising step $t$, we replace the original noise estimate $\epsilon_{\phi}(x_t, y)$ with a viewpoint-corrected version
\begin{equation}
\hat{\epsilon}_\phi(x_t, y)
= \epsilon_{\phi}(x_t, y) 
\;+\; \lambda_v \,\nabla_{x_t} E_{\text{vp}}(t),
\label{eq:eg_view}
\end{equation}
where $\lambda_v$ modulates the guidance strength. Since DDPM/DDIM updates subtract $\hat{\epsilon}_\phi$ from the state, adding $+\,\nabla_{x_t}E_{\text{vp}}(t)$ inside $\hat{\epsilon}_\phi$ results in a descent step on $E_{\text{vp}}(t)$ at the state level.


\paragraph{Adaptive guidance.}
In text-to-3D optimization, earlier timesteps mainly shape global geometry, while later timesteps refine appearance. To avoid over-constraining late steps, we adopt an adaptive schedule controlled by a no-reference IQA score $\mathcal{G}(x_0)$ (e.g., BRISQUE~\cite{mittal2012no}), where $x_0=g(\theta, v)$:
\begin{equation}
\lambda_v = \mathcal{F}\bigl(\mathcal{G}(x_0)\bigr),
\end{equation}
such that stronger structural guidance is applied early and gradually relaxed in later stages.

\begin{figure*}[htbp]
    \centering
    \includegraphics[width=1.0\linewidth]{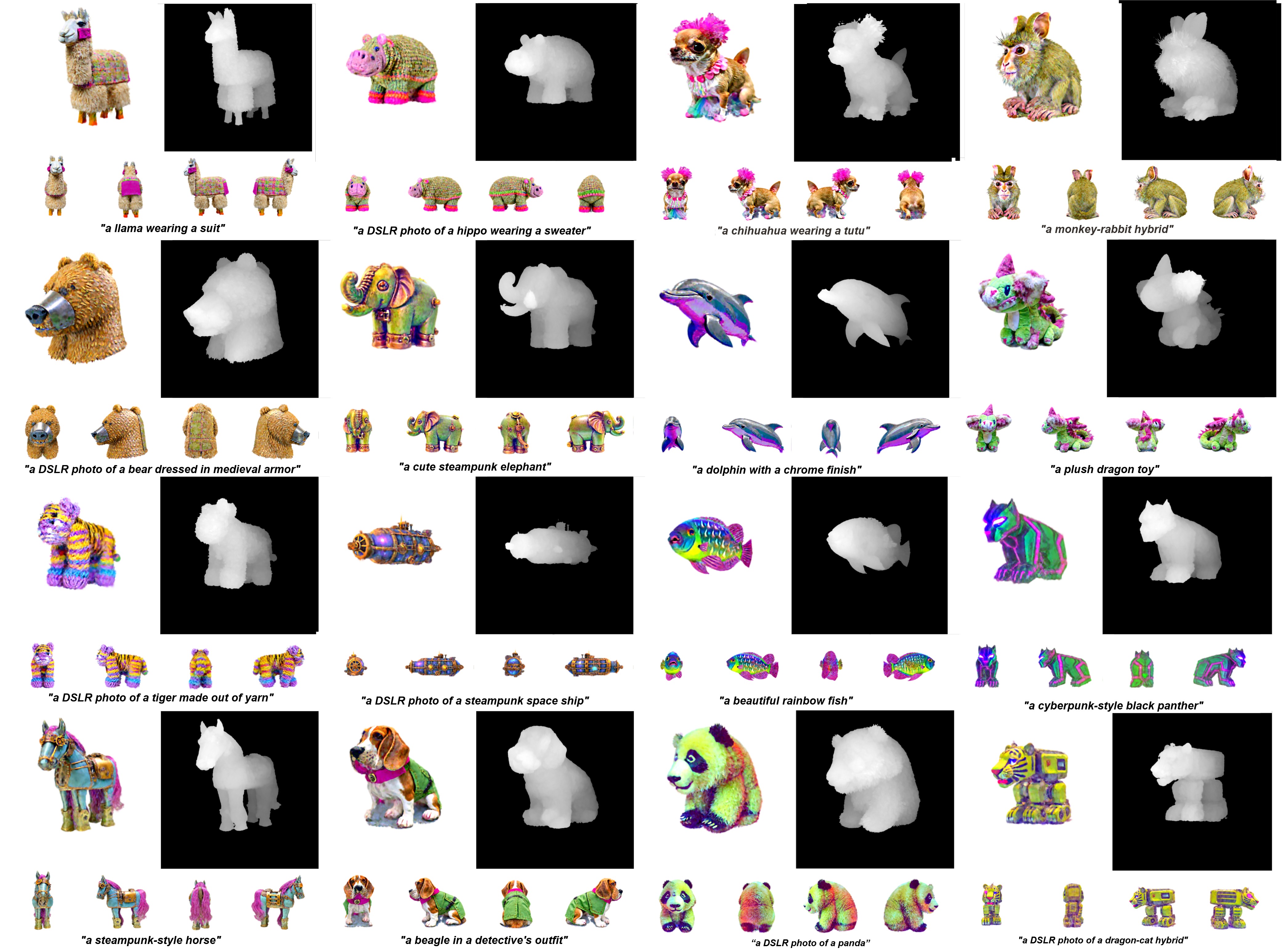}
    \caption{\textbf{Examples generated by SEGS.} SEGS can enhance the consistency of geometric shapes and the variety of high-fidelity textures in text-to-3D content generation. }
    \label{fig:main}
\end{figure*}

\subsection{Coupling with SDS/VSD}
\label{sec:coupling}

\noindent
The proposed guidance operates entirely at sampling time and integrates into standard text-to-3D pipelines as a plug-and-play module.
Substituting the viewpoint-corrected prediction \eqref{eq:eg_view} into the SDS objective (Eq.~\ref{equation1}) yields
\begin{equation}
\resizebox{0.97\linewidth}{!}{$
\nabla_{\theta} \mathcal{L}_{\text{SDS}}(\theta)
\;\approx\;
\mathbb{E}_{t,\epsilon,c}
\Bigl[
\,\omega(t)\,\bigl(\underbrace{\epsilon_{\phi}(x_t, y)
\;+\;\lambda_v\,\nabla_{x_t}E_{\text{vp}}(t)}_{\hat{\epsilon}_\phi(x_t, y)}
\;-\;\epsilon_t\bigr)\,\frac{\partial x_0}{\partial\theta}
\Bigr],
$}
\label{eq:combined}
\end{equation}
where an analogous replacement applies to VSD.
In practice, for each rendered view, we compute $E_{\text{vp}}(t)$ and its gradient once per denoising step, form $\hat{\epsilon}_\phi(x_t,y)$ as above, and feed it into the outer SDS (or VSD) optimization loop.
This preserves the original backbone and training-free property while imparting explicit viewpoint-aware structural control.

\paragraph{Text Consistency Guard }
A lightweight CLIP-based gate can optionally prune misaligned pseudo-supervision for the target-view text; thresholds and scheduling are deferred to the supplement.

\begin{figure*}[t]
    \centering
    \includegraphics[width=0.9\linewidth]{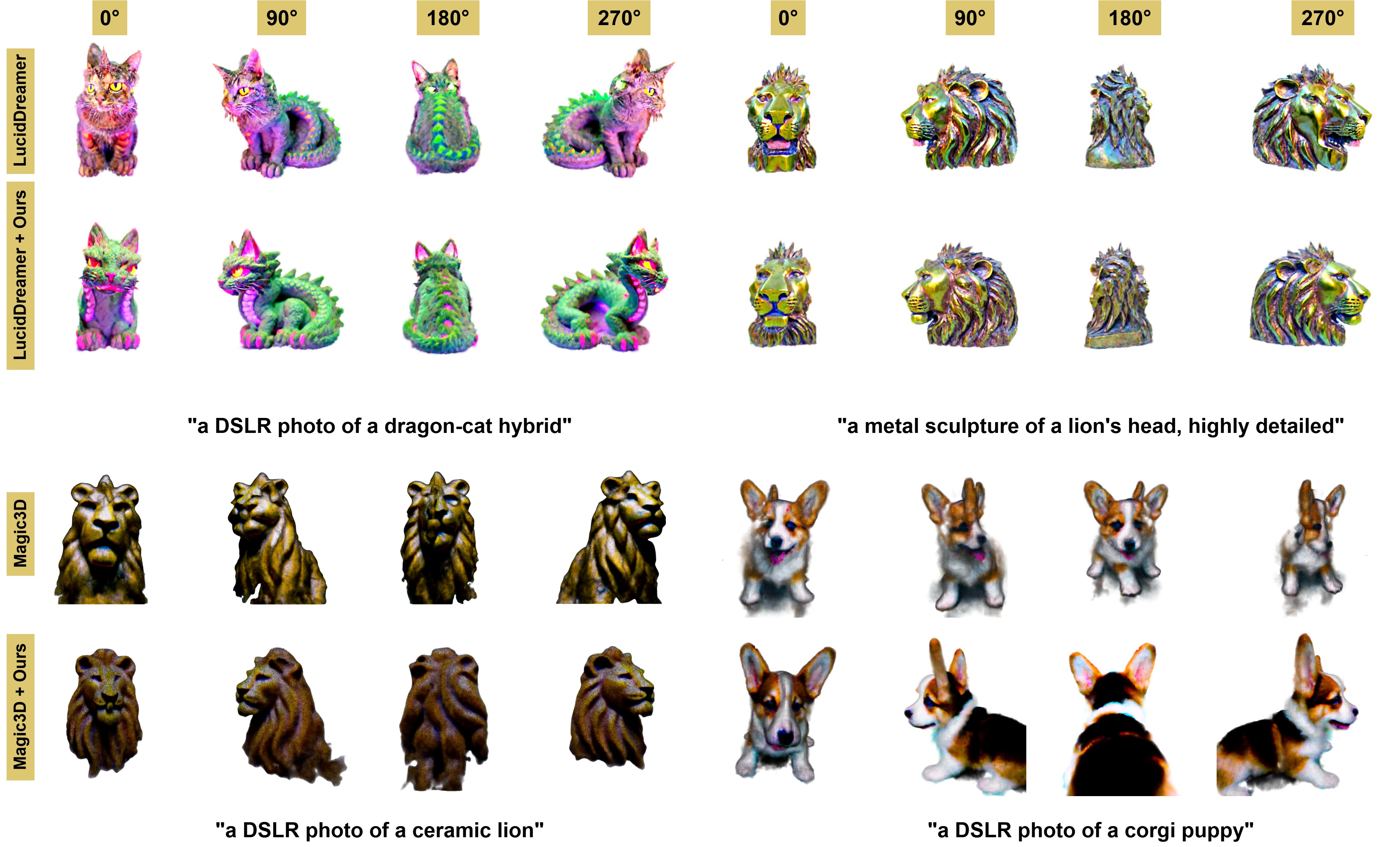}
    \caption{\textbf{Comparison with baseline methods in text-to-3D generation.} Adding our method to the baseline significantly improves the geometric consistency of the generation of the baseline. SEGS effectively mitigates the Janus problem.}
    \label{fig:example}
\end{figure*}
\section{Experiments}
\label{sec:exp}
We evaluate the proposed SEGS on the DreamFusion prompt library~\cite{poole2022dreamfusion}.
Both qualitative and quantitative comparisons are provided, showing improved \emph{view consistency}.
We also conduct ablations to assess the contribution of each component within SEGS.

\subsection{Implementation Details}
\label{sec:exp_implementation}

\noindent \textbf{Structure Feature Guidance.}
Unless otherwise stated, we extract self-attention \emph{keys} from the final layer of the first decoder up\_block of the U\mbox{-}Net and apply channel-wise mean centering before PCA.
To specify underrepresented viewpoints, we partition azimuth into three coarse bins: front ($[-60^\circ,60^\circ)$), side ($[-120^\circ,-60^\circ)\cup[60^\circ,120^\circ)$), and back ($[-180^\circ,-120^\circ)\cup[120^\circ,180^\circ)$), and focus guidance on the \emph{back} bin.
We generate auxiliary images with the prompt appended by ``back view'', compute CLIP similarities to the target-view text, and select top-$k$ samples to build a viewpoint-aware PCA basis $\mathbf{B}_t$ and references $\mathbf{S}_{g,t}$ (Sec.~\ref{sec:struct_pca}).
At each denoising step $t$, we project the current intermediate feature $\mathbf{f}$ onto $\mathbf{B}_t$ to obtain $\mathbf{G}_t$, compute the structural energy $E_{\text{vp}}(t)$ (Sec.~\ref{sec:struct_energy}), and inject its gradient into the noise prediction (Sec.~\ref{sec:eg_update}); the resulting $\hat{\epsilon}_\phi$ is then used in SDS/VSD via Eq.~\eqref{eq:combined}.
During optimization, $\mathbf{B}_t$ and $\mathbf{S}_{g,t}$ are treated as constants (no gradient through the PCA pipeline); gradients flow to $x_t$ through $\mathbf{f}$.

\noindent \textbf{Text Consistency Guard}
To filter clearly misaligned pseudo-supervision, we apply a CLIP-based gate comparing the decoded $\hat{x}_0^t$ with the target-view text (e.g., ``back view'').
Samples below an adaptive cosine-similarity threshold $\tau$ are discarded.

\noindent \textbf{Prompt Simplification.}  
To obtain object-centric supervision, we simplify complex prompts using GPT-4o. For each prompt from the DreamFusion library, the main object token is extracted and used both for generating auxiliary features and initializing 3D geometry. This improves consistency across structure-aware guidance and semantic pruning.

\subsection{Text-to-3D Generation}

\noindent \textbf{Protocol.}
Unless otherwise noted, we follow the setup in Sec.~\ref{sec:exp_implementation}.
For this experiment, we generate 20 auxiliary ``back view'' images per prompt with Stable Diffusion v1.4, set the PCA dimension to $N_b{=}64$ and CLIP selection to $k{=}3$, and run all methods on a single NVIDIA A100.
We use the simplified object token (Sec.~\ref{sec:exp_implementation}) to form the view-specific text templates used for auxiliary generation and evaluation.

\noindent \textbf{Qualitative Comparison.}
We primarily compare SEGS with representative text-to-3D baselines, including a vanilla SDS implementation via ThreeStudio~\cite{liu2023threestudio} and LucidDreamer~\cite{liang2024luciddreamer}. 
All methods distill from Stable Diffusion v1.4 under the same compute. 
We focus on SDS-style pipelines since SEGS is designed as a \emph{training-free, plug-and-play} module that directly enhances such frameworks; fine-tuned paradigms like Zero-1-to-3 or MVDream rely on additional curated data and retraining, which falls outside our scope. 
As shown in Fig.~\ref{fig:example}, SEGS visibly reduces Janus artefacts and improves view consistency relative to the baselines.
\begin{table}[t]
\centering

\begin{tabular}{lcc}
\toprule
\textbf{Method} & \textbf{JR (\%)$\downarrow$} & \textbf{View-CS $\uparrow$} \\
\midrule
LucidDreamer~\cite{liang2024luciddreamer} & 58.06 & 30.16 \\
LucidDreamer+SEGS (Ours) & \textbf{48.39} & \textbf{30.95} \\
\midrule
Magic3D~\cite{lin2023magic3d} & 68.18 & 32.17 \\
Magic3D+SEGS (Ours) & \textbf{56.36} & \textbf{32.85} \\
\midrule
DreamFusion~\cite{poole2022dreamfusion} & 72.73 & 32.85 \\
DreamFusion+SEGS (Ours) & \textbf{63.64} & \textbf{33.29} \\
\bottomrule
\end{tabular}
\label{table1}
\caption{Comparison of geometric consistency between baseline methods and SEGS-enhanced versions. Our method achieves a lower JR than all baselines, indicating better geometric consistency. It also improves View-CS, showing better alignment with the target viewpoints.}
\end{table}

\noindent \textbf{Quantitative Comparison.}
Previous research in 3D generation has lacked standardized evaluation metrics. Most existing methods rely on 2D image-based metrics to evaluate the visual quality or prompt relevance of generated content~\cite{heusel2017gans, hessel2021clipscore}. However, these metrics are inadequate for capturing viewpoint-dependent inconsistencies, particularly those characteristic of the Janus problem.

To more effectively evaluate geometric consistency across views, we introduce the Janus Problem Rate (JR). Given 124 text prompts, we generate 3D assets and render each object from four uniformly distributed azimuth angles. We then manually inspect the rendered views to identify geometric inconsistencies—such as missing or duplicated structures visible only from specific angles (e.g., back or side views). The total count of such failures across all prompts is reported as JR.

To introduce additional objectivity into the evaluation, we further propose a viewpoint-aware CLIP Score (View-CS). Specifically, we compare rendered images at azimuth angles of $0^\circ$, $90^\circ$, and $180^\circ$ against textual prompts such as \textit{"front view of an object"}, \textit{"side view of an object"}, and \textit{"back view of an object"}, respectively. This metric captures the alignment between the rendered viewpoint and the intended camera direction. Results show that our method achieves notable improvements in View-CS across all three baseline methods, while also significantly reducing JR, suggesting enhanced viewpoint fidelity without compromising geometric consistency.

\subsection{Ablation Study}
To verify the effectiveness of each component in our method, we carry out ablation studies.

\begin{figure}[t]
    \centering
    \includegraphics[width=1\linewidth]{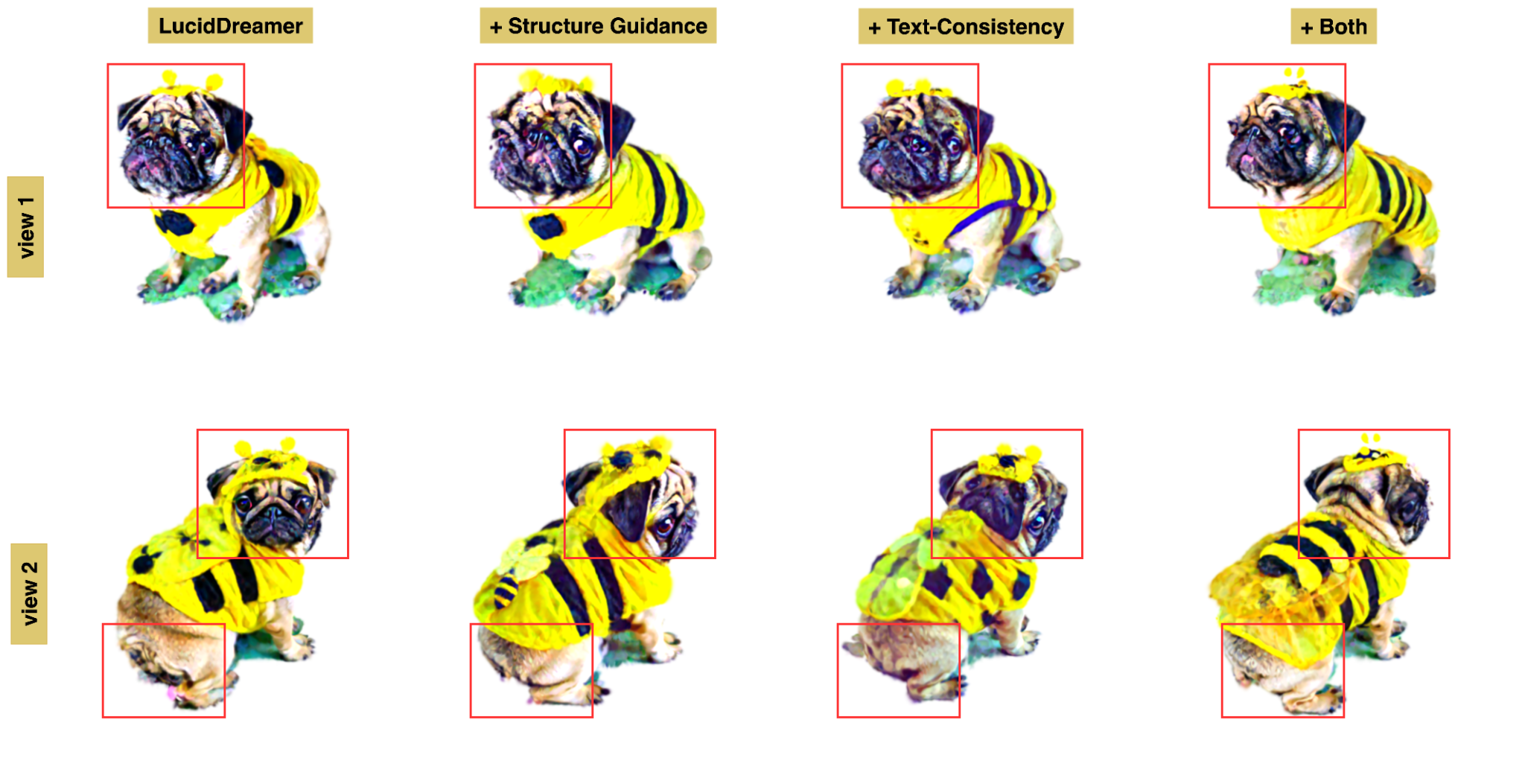}
     \caption{Ablation study on the effect of our guidance modules. We adopt LucidDreamer as the baseline. As illustrated in the figure, when the rendering viewpoint is set to the back view, incorporating \textit{SEGS} effectively reduces the presence of front-facing features. However, certain side-view artifacts, such as slight "head-turning," may still occur. The \textit{Guard} alone is less effective in suppressing undesired front-view elements. When combining \textit{SEGS + Guard}, the generation aligns more closely with the intended back-view perspective.}
    
    \label{fig:effect_guidance}
\end{figure}

\noindent \textbf{Effect of Guidance.} 
As shown in Figure~\ref{fig:effect_guidance}, we visualize renders at a fixed back-view camera to compare guidance variants. In both the baseline and the \emph{Text-Consistency Guard}-only settings, front-facing facial cues still leak into back views, revealing a strong front-view bias. Applying \emph{structural energy guidance} suppresses these artifacts but may leave mild geometric inconsistencies around the head region. In contrast, combining structural guidance with the text-consistency guard removes the residual leaks and yields more consistent, viewpoint-accurate 3D representations.

\begin{figure}[t]
    \centering
    \includegraphics[width=.9\linewidth]{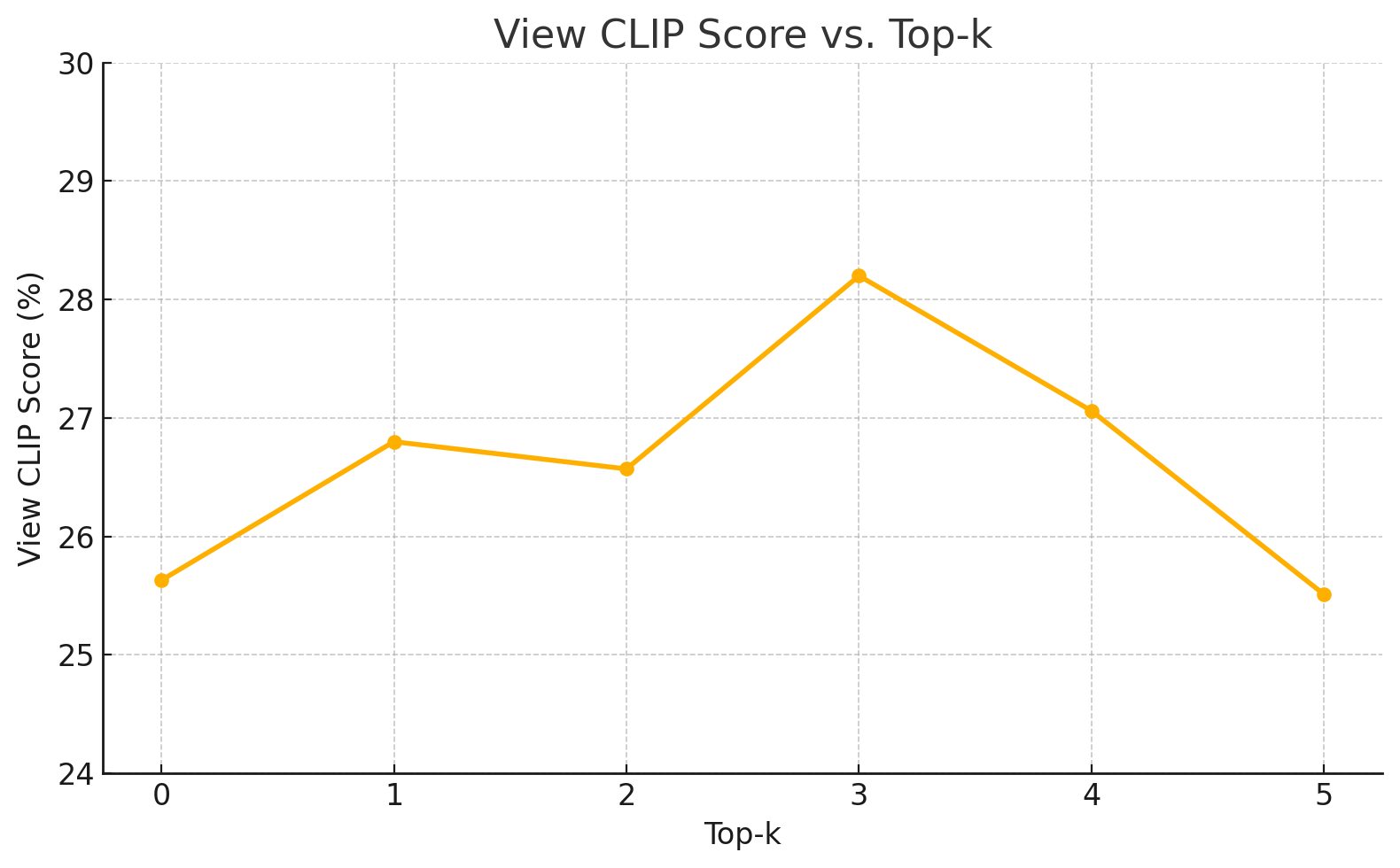}
    \caption{This figure illustrates how varying the number of selected images (Top-k) affects multi-view consistency, measured by View-CLIP Score.}
    \label{fig:top-k}
\end{figure}

\begin{figure}[t]
    \centering
    \includegraphics[width=1\linewidth]{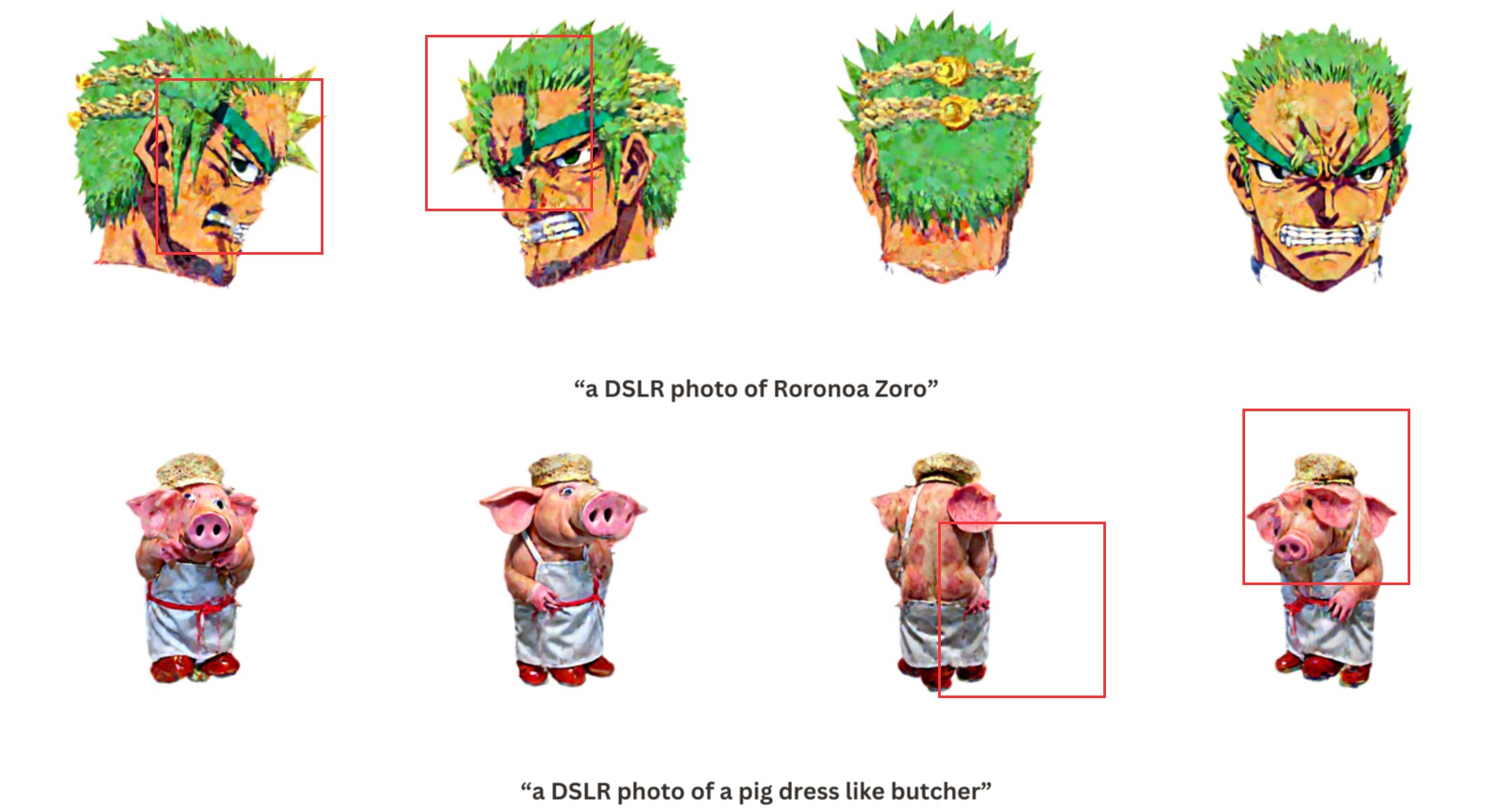}
    \caption{Failure cases at front--side transitions. \textbf{Red boxes} highlight residual frontal cues (e.g., eyes, ears, snout) that persist around lateral contours. These errors likely stem from ambiguous side-view evidence in the 2D prior and \emph{weaker structural targets at intermediate azimuths}, as our reference set focuses on the back.}
    \label{fig:failure_case2} 
\end{figure}

\begin{table}[t]
\centering
\begin{tabular}{lcc}
\toprule
\textbf{Method} & \textbf{SD1.4} & \textbf{SD2.1} \\
\midrule
DreamFusion & 100\% & 60\% \\
DreamFusion + SEGS (Ours) & \textbf{40\%} & \textbf{20\%} \\
\bottomrule
\end{tabular}
\caption{Janus Rate (JR↓) on DreamFusion with/without SEGS for SD-1.4 and SD-2.1 (single prompt, 5 seeds). SEGS consistently reduces JR by ~2.5× on both backbones.}

\label{tab:sd_versions}
\end{table}



\noindent \textbf{Top-k.}
Top-K refers to the number of images used to construct the guidance features after generating a feature subspace from the prompt. We evaluate its impact using the View CLIP Score (view-CS) across the front, side, and back rendered views of the generated 3D objects. As shown in Figure~\ref{fig:top-k}, we can see that as Top-K increases, the view-CS initially rises and then drops significantly. The highest score is observed at Top-3, indicating it is the most effective Top-K setting in our experiment.

\noindent \textbf{Comparison Across Diffusion Versions.}  
To examine how backbone models influence the Janus problem, we test both Stable Diffusion v1.4 and v2.1 using DreamFusion. The prompt “a DSLR photo of a corgi puppy” is randomly sampled five times. Table~\ref{tab:sd_versions} reports the average Janus Rate under each configuration.
While SD2.1 slightly reduces the frequency of Janus artifacts, both versions benefit substantially from our SEGS module. This validates the model-agnostic effectiveness of our training-free approach.


\noindent \textbf{Failure Case Analysis.}
While SEGS substantially reduces multi-face (Janus) artefacts, failures still occur near front--side transition angles (Fig.~\ref{fig:failure_case2}), where faint frontal cues may remain along lateral contours.
We attribute this to (i) ambiguity in side-view evidence in the 2D prior and (ii) weaker structural targets at intermediate azimuths because the current reference set concentrates on the back bin.
\section{Conclusion}
\label{sec:conclusion}
In this paper, we show that viewpoint bias in the training images of diffusion models is a key factor underlying the Janus problem in text-to-3D generation. 
To address this issue, we propose SEGS, a training-free framework that provides hierarchical guidance by leveraging structure-based features from intermediate U-Net activations and a lightweight CLIP-based semantic guard. 
SEGS effectively reduces multi-head artifacts while preserving appearance fidelity, and can be seamlessly integrated as a plug-and-play module into existing text-to-3D pipelines. 
Looking forward, we observe that finer-grained cues such as facial details tend to emerge in later U-Net layers; exploring guidance that explicitly steers trajectories away from these late-stage frontal features may offer a promising direction for further mitigating Janus artifacts.

{
    \small
    \bibliographystyle{ieeenat_fullname}
    \bibliography{main}
}

\end{document}